\documentclass[runningheads]{llncs}
\usepackage{amsmath}
\usepackage{amssymb}
\usepackage{colortbl}
\usepackage[T1]{fontenc}
\usepackage{graphicx,verbatim}
\usepackage{xcolor}
\usepackage{booktabs}
\usepackage{multirow}

\usepackage{tikz}
\usepackage{colortbl}
\usepackage{hyperref}
\newcommand{\Ev}[1]{\hyperref[ev:#1]{[A#1]}}
\usetikzlibrary{shadows}
\definecolor{kleinblue}{RGB}{0, 47, 167} 
\definecolor{kleinblue2}{RGB}{20, 20, 125} 

\newcommand*\circled[1]{%
\tikz[baseline=(char.base)]{
  \node[shape=circle, fill=kleinblue2, draw=white, thick, drop shadow={shadow xshift=0.2ex,shadow yshift=-0.2ex}, inner sep=1pt] (char) {\textcolor{white}{#1}};
}}
\begin{document}
\title{ClinCoT: Clinical-Aware Visual Chain-of-Thought for Medical Vision Language Models}
\titlerunning{ClinCoT: Clinical-Aware Visual Chain-of-Thought for Med-VLMs}

\author{Xiwei Liu\inst{1} \and
Yulong Li\inst{1} \and
Xinlin Zhuang\inst{1,2} \and
Xuhui Li\inst{1}
\and
Jianxu Chen\inst{1} \and
Haolin Yang\inst{1} \and
Imran Razzak\inst{1} \and
Yutong Xie\inst{1,}$^{\dagger}$}
\authorrunning{X. Liu et al.}
%
\institute{Mohamed bin Zayed University of Artificial Intelligence \and
East China Normal University  \\
$^{\dagger}$Corresponding Author\\
}

\maketitle              
\begin{abstract}
Medical Vision-Language Models have shown promising potential in clinical decision support, yet they remain prone to factual hallucinations due to insufficient grounding in localized pathological evidence. Existing medical alignment methods primarily operate at the response level through preference optimization, improving output correctness but leaving intermediate reasoning weakly connected to visual regions. 
%
Although chain-of-thought (CoT) enhances multimodal reasoning, it remains largely text-centric, limiting effective integration of clinical visual cues.
To address this gap, we propose ClinCoT, a clinical-aware visual chain-of-thought framework that transforms preference optimization from response-level correction to visual-driven reasoning. We introduce an automatic data generation pipeline that constructs clinically grounded preference pairs through reasoning with hypotheses-driven region proposals. Multiple Med-LLMs evaluators rank and assign scores to each response, and these rankings serve as supervision to train the target model. 
We further introduce a scoring-based margin-aware optimization strategy that incorporates both preference ranking and score difference to refine region-level reasoning trajectories. To maintain alignment as the model's policy evolves during training, we adopt an iterative learning scheme that dynamically regenerates preference data. 
Extensive experiments on three medical VQA and report generation benchmarks demonstrate that ClinCoT consistently improves factual grounding and achieves superior performance compared with existing preference-based alignment methods. Our code is available in \url{https://github.com/Xiwei-web/ClinCoT}


\end{abstract}
\section{Introduction}
Medical Vision-Language Models (Med-VLMs) have demonstrated promising potential in assisting clinical decision-making, including medical visual question answering (Med-VQA) and radiology report generation~\cite{li2023llava}\cite{liu2024zero}\cite{liu2025fias}\cite{zhu2025sss}. However, despite recent advances, Med-VLMs still suffer from a fundamental limitation: insufficient alignment between visual evidence and generated clinical conclusions~\cite{zhu2024mmedpo}\cite{zhou2024aligning}. In particular, models with poor modality alignment may rely heavily on pretrained language priors while underutilizing localized pathological evidence, leading to hallucinated findings or clinically irrelevant responses~\cite{chen2024detecting}\cite{jiang2024medthink}.

To improve factual consistency, recent medical alignment methods adopt preference optimization strategies~\cite{amini2024direct}\cite{hein2024preference}. Clinical-aware preference learning frameworks such as MMedPO~\cite{zhu2024mmedpo} construct medically meaningful preference pairs and introduce weighted optimization objectives to emphasize clinically relevant responses. Other alignment approaches refine model outputs through self-rewarding mechanisms~\cite{yuan2024self}, self-training with stronger evaluators~\cite{sun2024stllava}, or modality alignment objectives~\cite{zhou2024aligning}. While these methods improve response-level factuality and reduce overt hallucination, they largely operate at the output level, treating each response as a monolithic entity. Consequently, they do not explicitly model how localized pathological regions influence intermediate reasoning steps. The lack of region-aware reasoning limits the interpretability and pathological grounding of Med-VLMs, especially in complex diagnostic scenarios.

Beyond preference optimization, Chain-of-thought (CoT) reasoning~\cite{wei2022chain}\cite{dai2025goca}\cite{wu2025rankcot} has emerged as an effective strategy for improving complex reasoning and step-wise decision-making in both language and vision-language models. In medical contexts, structured diagnostic prompting and reasoning-enhanced training strategies have been shown to improve explanation quality and logical consistency~\cite{singhal2023large}\cite{singhal2025toward}. However, most existing CoT approaches remain predominantly text-centric~\cite{dai2025goca}: they guide models to generate sequential reasoning tokens without explicitly restructuring visual attention. This implicitly assumes that the visual encoder uniformly captures all clinically relevant information, which is often unrealistic in medical imaging. Diagnostic reasoning in radiology fundamentally depends on detecting and examining localized abnormalities, such as small nodules, subtle consolidations, or focal fractures~\cite{xu2025lingshu}.

In medical practice, clinicians rarely reason over entire images uniformly. Instead, they form differential diagnostic hypotheses, examine disease-relevant regions, and iteratively refine conclusions based on localized evidence. This naturally raises a fundamental question:

\textit{\textcolor{kleinblue2}{Can preference optimization be extended beyond response-level correction toward hypotheses-driven clinical reasoning?}}

To address this challenge, we propose ClinCoT, a clinical-aware visual CoT framework that unifies region-level diagnostic hypotheses with margin-aware preference optimization under a coherent reasoning paradigm. Rather than merely improving final outputs, ClinCoT explicitly models how localized pathological evidence shapes intermediate reasoning trajectories, thereby encouraging closer alignment between visual grounding and clinical inference in a process-driven manner. Specifically, we design an automatic two-stage pipeline to construct clinically grounded preference data: 1) Hypotheses-Driven Region Generation: Given a medical image, we generate multiple disease-conditioned region using a clinical-aware visual tool. Then the target Med-VLM answers the question by jointly processing the original image and each candidate region, forming multiple pathology-aware reasoning chains. 2) Consensus-Weighted Quality Assessment: Multiple medical LLMs evaluators score the generated responses. These scores serve as proxies for the clinical validity of the region-conditioned reasoning chains and are adjusted based on evaluator agreement to ensure robust supervision.

\textbf{Our contributions} are three-fold: \circled{1} an automatic clinical hypotheses-driven pipeline for scalable region-level preference data construction; \circled{2}
a consensus weighted scoring based preference optimization with iterative learning for pathology-aware reasoning alignment, enabling finer discrimination of key regions and progressively stabilizing reasoning trajectories; and \circled{3}
extensive experiments on multiple medical VQA and report generation benchmarks demonstrating consistent improvements over strong medical baselines.

\begin{figure}[t]
    \centering
    \includegraphics[width=\linewidth]{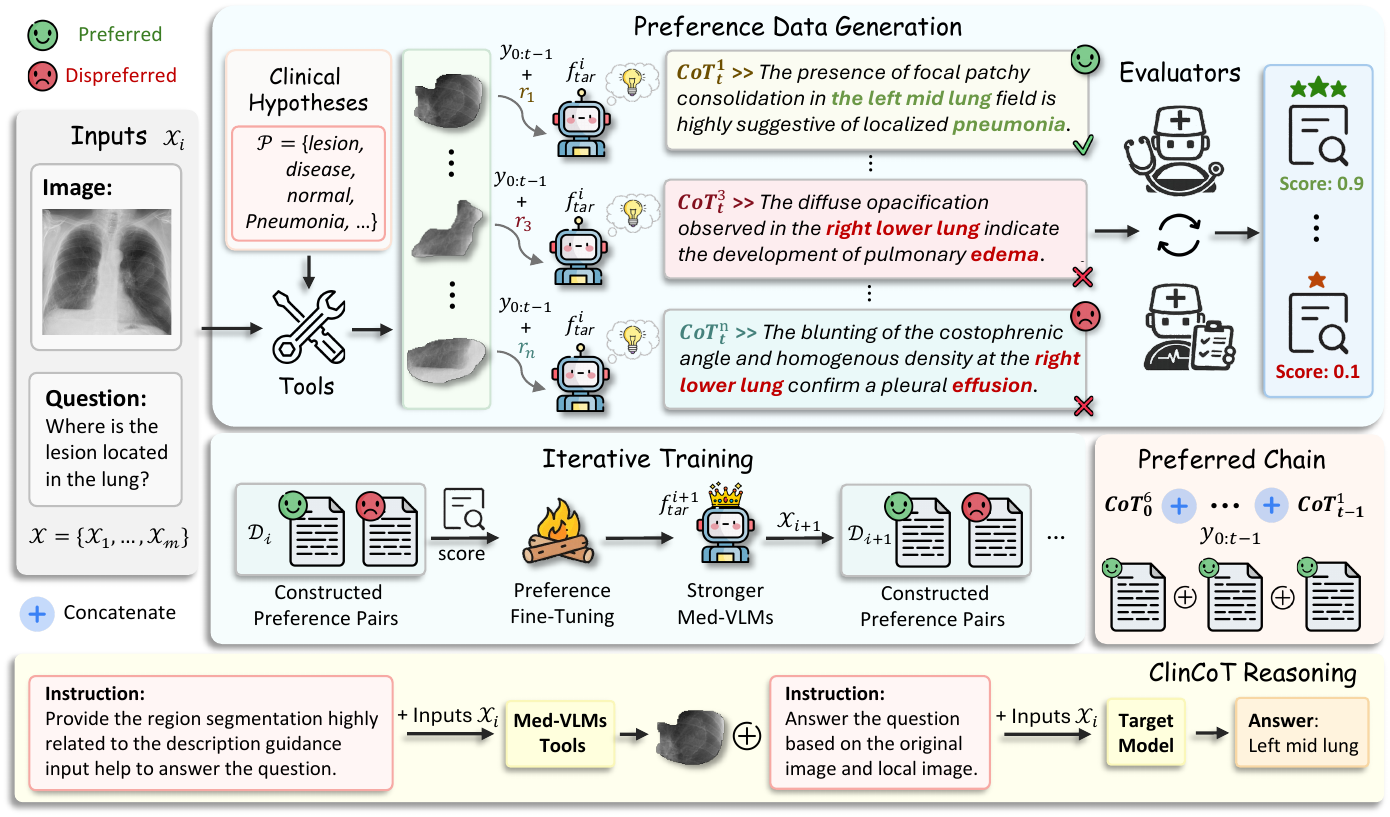}
    \caption{ \textbf{Overview of the ClinCoT workflow.} 
Given a subset of input pairs $\mathcal{X}_i \subset \{\mathcal{X}_1,\ldots,\mathcal{X}_m\}$, ClinCoT first employs a clinical-aware tool with a predefined hypotheses set $\mathcal{P}$ to generate region proposals $\{r_i\}_{i=1}^n$. 
The target model $f_{\mathsf{tar}}^i$ produces region-conditioned reasoning chains $\{y_t^i = CoT_t^i\}_{i=1}^n$ based on the preserved preferred history $y_{0:t-1}=\{CoT_0^6,\ldots,CoT_{t-1}^1\}$, integrating both the original image and the candidate regions. Med-LLM evaluators assign scores to construct preference pairs $\mathcal{D}_i$, distinguishing preferred and dispreferred responses. 
A consensus-weighted scoring based optimization updates $f_{\mathsf{tar}}^i$ to $f_{\mathsf{tar}}^{i+1}$ through iterative training. 
The updated model $f_{\mathsf{tar}}^{i+1}$ is then applied to $\mathcal{X}_{i+1}$ to generate new preference pairs $\mathcal{D}_{i+1}$ for the next iteration.
    }
    \label{fig:overview}
\end{figure}

\section{Methodology}
Given a target model $f_\mathsf{tar}$, a set of evaluator models $f_\mathsf{eval}$, and an image-question pair $x=\{x_v,x_q\}$, we illustrate how to construct $n$ preference data points, as shown in Figure~\ref{fig:overview}. 
Assuming a reasoning chain of $T$ steps, preference data are progressively generated throughout these $T$ stages.
At each timestep $t$, the pipeline consists of three stages: Response Generation, Response Evaluation and Pair Construction (see Sec.~\ref{sec:generation}). Unlike standard Direct Preference Optimization (DPO)~\cite{amini2024direct}, we adopt a consensus-weighted scoring based DPO, which not only ranks preference data but also assigns preference scores. This scoring enables more precise optimization based on score differences. During training, the rankings of the preference data act as supervision by minimizing negative log-likelihood loss, while the scores define the decision margin (see Sec.~\ref{sec:preference}).

\subsection{Preference Data Generation}
\label{sec:generation}
\noindent\textbf{Response Generation.} 
This stage aims to construct clinically grounded region hypotheses and generate intermediate responses conditioned on localized visual contexts. A `response' refers to any model output at an intermediate reasoning step, not necessarily the final answer to the question. We denote the response at timestep $t$ as $y_t$, with the initial input $x$ represented as $y_0$.
Given a medical image $x_v$ and a predefined set of clinical hypotheses $\mathcal{P} = \{p_i\}_{i=1}^n$, we employ a clinical-aware VLM (e.g., MedKLIP~\cite{wu2023medklip}) $\mathcal{T}(\cdot)$ to obtain disease-conditioned activation maps, which highlights image regions associated with the clinical concept $p_i$:
\begin{equation}
    h_i = \mathcal{T}(x_v, p_i), \quad i = 1, \dots, n,
\end{equation}
where each $h_i \in [0,1]^{H \times W}$ is then used to localize clinical region via thresholding and connected-component extraction: $r_i = x_v \odot \Phi(h_i)$. Here, $\Phi(\cdot)$ denotes a deterministic region extraction operator with $\odot$ representing element-wise masking. For each region hypothesis $r_i$, the target Med-VLM $f_{\mathsf{tar}}$ generates an intermediate response conditioned on both the global image and the localized region:
\begin{equation}
    y_t^i = f_{\mathsf{tar}}(x_v, r_i, x_q \mid y_{0:t-1}),
\end{equation}
resulting in a set of candidate responses $\{y_t^i\}_{i=1}^n$ that reflect correspond to different region-conditioned visual interpretations of the same image, where $y_{0:t-1}$ denotes the currently preserved reasoning chain from previous timesteps.

\noindent\textbf{Response Evaluation.} This stage evaluates the quality of all generated responses by prompting evaluator model to assign a score between 0 and 1, with higher scores indicating better prediction. The evaluator not only scores each individual response but also counts how this response influences the quality of its next response in the chain. This cumulative evaluation approach helps quantify the impact of each bounded region on the overall reasoning process. At timestep $t$, the evaluator assigns scores for $y_t^i$ as follows:
\begin{equation} 
s_{\mathsf{cur}}^i =f_\mathsf{eval}(y^i_t \mid y_{0:t-1}), \quad
s_{\mathsf{nxt}}^i =\mathbb{E}_j[f_\mathsf{eval}(y_{t+1}^j \mid y_{0:t-1}, y^i_t)], \quad
s^i =s_{\mathsf{cur}}^i+\gamma s_{\mathsf{nxt}}^i,
\end{equation} 
where $s_{\mathsf{nxt}}^i$ reflects the impact on the next response with expectation $\mathbb{E}[\cdot]$ estimated by randomly sampling $j$ next responses according to current response $y_t^i$. $\gamma>0$ is a hyperparameter to combine the current and next response scores, with $\gamma=0$ at the last step. To mitigate the inherent bias of a single Med-LLM and ensure reliable assessments, we employ a \textit{Consensus-Weighted Scoring} strategy utilizing two distinct evaluators. we calculate the agreement between the two evaluators to penalize controversial assessments as follow:
\begin{equation}
    s^i_\mathsf{{final}} = \left( \frac{s_1^i + s_2^i}{2} \right) \cdot \exp\left( -|s_1^i - s_2^i| \right)
\end{equation}
where $s_1^i$ and $s_2^i$ are scores given by two Med-LLM evaluators respectively.

\noindent\textbf{Pair Construction.} From these scored reasoning response candidates $\{y_t^i\}_{i=1}^n$ with $\{s^i_\mathsf{{final}}\}_{i=1}^n$, we select $k$ pairs to construct preference comparisons at each timestep $t$. For each pair, the response with the higher score is treated as the preferred response and concatenated with the historical chain $y_{0:t-1}$ to form the ``preferred chain'' $y_t^w$. Similarly, the lower-scoring response is concatenated with the same historical chain to form the ``dis-preferred chain'' $y_t^l$. Each pair of chains, together with their corresponding scores $(y_w, s_w, y_l, s_l)$, constitutes one preference training example. The collection of all $k$ such pairs forms the preference dataset $\mathcal{D}_t$ for timestep $t$. To ensure a stable forward reasoning trajectory, only the highest-scoring response at timestep $t$ is concatenated with the historical chain to form the updated reasoning chain $y_{0:t}$.
This updated chain is then used as the input for the next timestep. 
In other words, although multiple preference pairs are created for learning purposes at each timestep, only the single best-scoring chain is retained to continue the forward reasoning process.

\subsection{Preference Fine-tuning}
\label{sec:preference}
\noindent\textbf{Margin-Aware Optimization.} Give an input $x$, a language model policy $\pi_\theta$ can produce a conditional distribution $\pi_\theta(y\mid x)$ with $y$ as the output text response. Standard DPO~\cite{amini2024direct} reformulates reward learning as policy optimization by reparameterizing the reward under the PPO framework~\cite{schulman2017proximal}:
\begin{equation}
\small
r(x,y)=\beta \log \frac{\pi_\theta(y\mid x)}{\pi_{\mathrm{ref}}(y\mid x)}+\beta \log Z(x).
\end{equation}
Here $\pi_{\mathrm{ref}}$ denotes the frozen initialization of $\pi_\theta$ controlling the policy regularization, $\beta$ is a temperature parameter and $Z(x)$ is a $y$-independent normalization term. 
Under the Bradley–Terry model~\cite{bradley1952rank}, 
DPO minimizes the negative log-likelihood that a preferred response $y_w$ is ranked above the dispreferred one $y_l$:
\begin{equation}
\small
\begin{aligned}
&P(y_w \succ y_l)=\sigma\!\left(r(x,y_w)-r(x,y_l)\right), \\
&\mathcal{L}_{\mathrm{DPO}}=-\mathbb{E}_{(x,y_w,y_l)\sim\mathcal{D}}
\left[\log P(y_w \succ y_l)\right],
\end{aligned}
\end{equation}
where $\sigma(\cdot)$ denotes the sigmoid function. In image-level reasoning, different regions contribute unequally to the final decision, motivating finer discrimination between preference pairs. 
To capture this variability, we introduce a margin term derived from preference scores and formulate a margin-aware objective:
\begin{equation}
\label{eq:dpo_ours}
\scriptsize
\mathcal{L}_{\mathrm{ClinCoT}}(\theta)
=
-\mathbb{E}_{(x,y_w,y_l)\sim\mathcal{D}}
\left[
\log \sigma
\left(
\beta \log \frac{\pi_\theta(y_w\mid x)}{\pi_{\mathrm{ref}}(y_w\mid x)}
-
\beta \log \frac{\pi_\theta(y_l\mid x)}{\pi_{\mathrm{ref}}(y_l\mid x)}
-
(g(s_w)-g(s_l))
\right)
\right].
\end{equation}
Here, the function $g(\cdot)$ is monotonically increasing and maps preference scores into the logit space of the DPO objective, amplifying the key region’s influence.

Let $\Delta_{r}=g(s_w)-g(s_l)$ and define the Gumbel-distributed random variables $R_w \sim \operatorname{Gumbel}\left(r\left(x, y_w\right), 1\right)$ and $R_l \sim \operatorname{Gumbel} \left(r\left(x, y_l\right), 1\right)$. The probability that the winning response is preferred, adjusted by the preference gap, is:
\begin{equation} 
\small
\begin{aligned}
&P\left(R_w-R_l>\Delta_{r}\right) =\sigma\left(r\left(x, y_w\right)-r\left(x, y_l\right) - \Delta_{r}\right) \\
& =\sigma\left(\beta \log \frac{\pi_\theta\left(y_w \mid x\right)}{\pi_{\mathrm{ref}}\left(y_w \mid x\right)}-\beta \log \frac{\pi_\theta\left(y_l \mid x\right)}{\pi_{\mathrm{ref}}\left(y_l \mid x\right)}- \Delta_{r}\right).
\end{aligned}
\end{equation} 
This result follows the definition of Gumbel random variables \cite{amini2024direct} and the Gumbel-max trick \cite{Maddison2017Gumbel}, with a similar derivation found in ODTO~\cite{amini2024direct}.
The Gumbel distribution models the extreme values of a variable, while $\Delta_{r}$ quantifies the degree of preference pairs difference. By maximizing the log-likelihood, we then obtain the adjusted loss function which explicitly optimizes preference learning by distinguishing not only the order but also the magnitude of preference differences.

\noindent\textbf{Iterative Learning.}
Standard DPO optimizes on a static preference dataset, which may induce distributional mismatch as the model evolves during training. 
To alleviate this issue, we adopt an iterative preference learning strategy inspired by~\cite{yu2024rlaif}, where \texttt{ClinCoT} refines the target model through incrementally updated data generation and optimization.
Specifically, the full image-question set $\mathcal{X}$ is partitioned into $m$ disjoint subsets, $\mathcal{X}=\{\mathcal{X}_1,\ldots,\mathcal{X}_m\}$, each corresponding to one iteration.
Starting from an initial target model $f_\mathsf{tar}$ and evaluators $f_\mathsf{eval}$, iteration $i$ first uses the current model $f^i_\mathsf{tar}$ to generate preference data $\mathcal{D}_i$ on subset $\mathcal{X}_i$.
The model is then updated to $f^{i+1}_\mathsf{tar}$ by optimizing Eq.~(\ref{eq:dpo_ours}) on $\mathcal{D}_i$.
This procedure repeats for $m$ rounds, yielding the final model $f^m_\mathsf{tar}$. 

\begin{figure}[t]
    \centering
    \includegraphics[width=\linewidth]{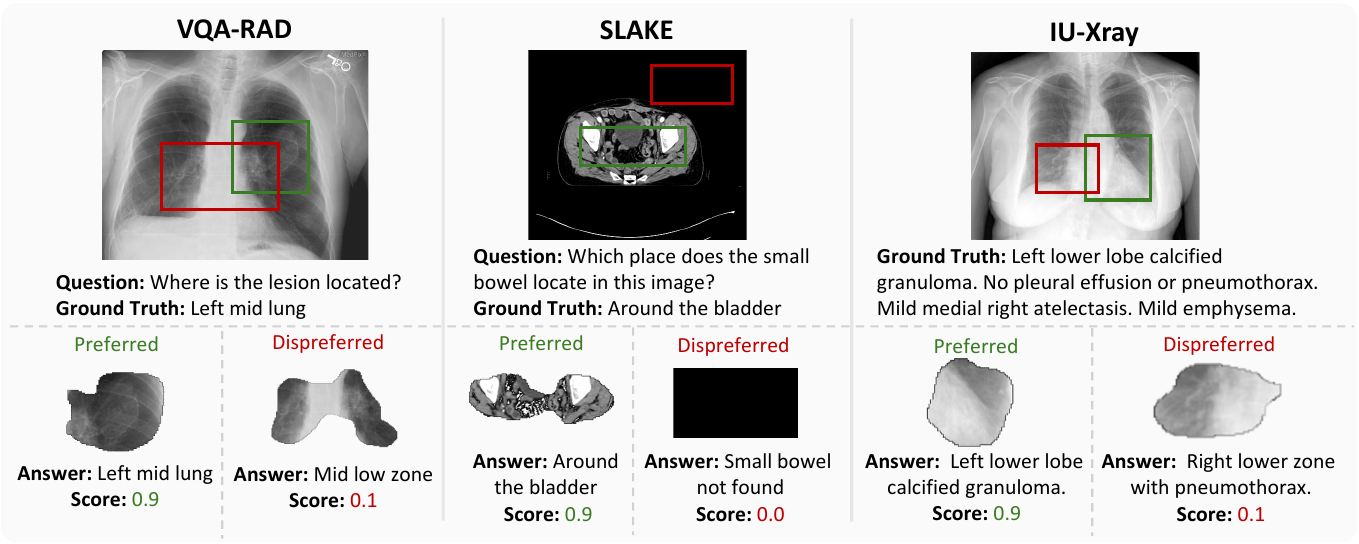}
    \caption{Visualization of generated preference data.}
    \label{fig:preference data}
\end{figure}

\section{Experiments}
\subsection{Experiment Setup}
\noindent\textbf{Evaluation Datasets.}
To verify the effectiveness of \texttt{ClinCoT} in improving factuality, we utilize three medical benchmarks: two medical VQA datasets, i.e., VQA-RAD~\cite{lau2018dataset} and SLAKE~\cite{liu2021slake}, and one report generation datasets IU-Xray~\cite{demner2016preparing}. 

\noindent
\textbf{Implementation Details.}
We utilize LLaVA-Med-1.5 7B~\cite{li2023llava} as the target model and apply LoRA fine-tuning~\cite{hu2021lora} for preference optimization with a batch size of 4, a learning rate of 1e-7, $\gamma=0.3$, $\beta=0.1$ and train for 3 epochs. Each reasoning chain contains $T=3$ timesteps, with $k=2$ per step. Iterative learning of ClinCoT is performed over $m=4$ rounds. LLaMA3-Med42-7B~\cite{christophe2024med42} and BioMistral-7B~\cite{labrak2024biomistral} are used as Med-LLM evaluators. All experiments are implemented using PyTorch 2.1.2 on two NVIDIA RTX A100 GPUs.

\begin{table*}[t]
\centering
\caption{Performance comparison on medical VQA and report generation tasks. For open-ended questions, we report recall (Open); for closed-ended questions, accuracy (Closed). The BLEU score denotes the average of BLEU-1/2/3/4. +SFT indicates that the model is first fine-tuned with SFT before applying the corresponding baselines. The best results and second best results are bold and \underline{underlined}, respectively.}
\resizebox{\linewidth}{!}{
\begin{tabular}{l|cc|cc|ccc}
\toprule
\multirow{2}{*}{Models} & \multicolumn{2}{c|}{\textbf{SLAKE}} & \multicolumn{2}{c|}{\textbf{VQA-RAD}} & \multicolumn{3}{c}{\textbf{IU-Xray}}  \\ 
 & Open & Closed & Open  & Closed & BLEU  & ROUGE-L & METEOR   \\ \midrule
LLaVA-Med v1.5          & 44.26  & 61.30         & 29.24  & 63.97         & 14.56 & 10.31 & 10.95 \\ \midrule
+ DPO   & 49.30   & 62.02        & 29.76  & 64.70   & 16.08 & 12.95 & 17.13  \\ 
+ Self-Rewarding & 42.63 &  61.30 &  {33.29} & 64.17 & 14.20 & 10.38 & 10.52  \\
+ STLLaVA-Med & 48.65 & 61.75 & 30.17 & 64.38 & 16.11 & 10.58 & 10.51  \\ 
+ POVID & 52.43 & 70.35 & 31.77 &  {65.07} & 20.80 & 24.33 & 30.05 \\
+ SIMA & 51.77 & 69.10 & 31.23 & 64.80 & 17.11 & 22.87 & 29.10  \\
+ FiSAO &  {52.69} &  {70.46} & 32.70 & 64.11 &  {21.06} &  {25.72} &  {30.82}  \\
+ MMedPO 
&  \textbf{53.99} 
&  \underline{73.08}  
&  \textbf{36.36}  
&  \textbf{66.54}   
&  \underline{23.49}      
&  \underline{29.52}       
&  \underline{34.16}     \\
\rowcolor[gray]{0.9}+ \textbf{ClinCoT}  &  \underline{53.73}\scriptsize{$\pm0.3$} &  \textbf{73.41}\scriptsize{$\pm0.2$} &  \underline{33.15}\scriptsize{$\pm0.2$} &  \underline{65.88}\scriptsize{$\pm0.1$} &  \textbf{24.96}\scriptsize{$\pm0.2$} &  \textbf{29.79}\scriptsize{$\pm0.1$} & \textbf{35.44}\scriptsize{$\pm0.2$} \\
\midrule
+ SFT & 50.45 & 65.62 & 31.38 & 64.26 & 22.75& 28.86 & 33.66  \\
\quad + DPO  &  {53.50} & 69.47 & 32.88 & 64.33 & 23.07 &  {29.97} & 34.89 \\
\quad + Self-Rewarding & 50.62 & 65.89 & 32.69 &  {65.89} & 22.89 & 28.97 & 33.93  \\
\quad + STLLaVA-Med & 52.72 & 66.69 &  {33.72} & 64.70 & 22.79 & 28.98 & 34.05  \\
\quad + POVID &  52.18 & 70.67 & 32.95 & 64.97 & {23.95} & 29.75 & 34.63 \\
\quad + SIMA & 51.75 & 69.28 & 32.50 & 64.08 &23.90 & 29.41 & 34.45  \\
\quad + FiSAO & 52.80 &  {70.82} & 32.94 & 65.77 & 23.57 & 29.88 &  {35.01}  \\
\quad + MMedPO  &  \underline{55.23} &  \underline{75.24} &  \underline{34.03} &  \underline{67.64} &  \underline{24.00} &  \underline{30.13} & \underline{35.17} \\
\rowcolor[gray]{0.9}\quad + \textbf{ClinCoT}  &  \textbf{56.13}\scriptsize{$\pm0.5$} &  \textbf{75.77}\scriptsize{$\pm0.2$} &  \textbf{34.21}\scriptsize{$\pm0.1$} &  \textbf{67.89}\scriptsize{$\pm0.4$} &  \textbf{26.17}\scriptsize{$\pm0.2$} &  \textbf{31.73}\scriptsize{$\pm0.2$} & \textbf{36.59}\scriptsize{$\pm0.1$}  \\
\bottomrule 
\end{tabular}
}
\label{tab:results}
\end{table*}

\noindent
\textbf{Baselines.} 
We compare \texttt{ClinCoT} against standard DPO~\cite{rafailov2023direct} and its variants, including the self-rewarding approach~\cite{yuanself} and STLLaVA-Med~\cite{sun2024stllava}. The self-rewarding method constructs preference pairs from model-generated responses, while STLLaVA-Med enhances preference selection via GPT-4o for medical alignment. We also evaluate three VLM preference fine-tuning methods originally developed for natural images: POVID~\cite{zhou2024aligning}, SIMA~\cite{wang2024enhancing}, and FiSAO~\cite{cui2024fine}, as well as the medical-specific MMedPO~\cite{zhu2024mmedpo}. To further evaluate training compatibility, we assess \texttt{ClinCoT} and all baselines under supervised fine-tuning (SFT) settings to examine performance gains across training paradigms.

\noindent
\textbf{Evaluation Metrics.}
For Med-VQA task, we use accuracy and recall for both closed-ended and open-ended questions. For the report generation task, we use BLEU Score~\cite{papineni2002bleu}, ROUGE-L~\cite{lin2004rouge} and METEOR~\cite{banerjee2005meteor} as the metrics.

\subsection{Main Results}
The overall performance comparisons are reported in Table.~\ref{tab:results}. \texttt{ClinCoT} achieves the strongest performance among the compared baselines on report generation, but it underperforms MMedPO on VQA-RAD. This may suggest that response-level clinical weighting remains competitive for short-form VQA answers, where concise linguistic precision dominates. In contrast, \texttt{ClinCoT} emphasizes region-conditioned reasoning chains; without prior task adaptation, intermediate reasoning steps may introduce instability. Under SFT-enhanced setting, \texttt{ClinCoT} delivers the strongest overall performance. This suggests that SFT may provide a more stable domain-aligned initialization, which facilitates subsequent hypotheses-driven refinement. Once the base model is aligned with domain style, \texttt{ClinCoT} refines both answers and reasoning trajectories, yielding more consistent gains. Figure~\ref{fig:preference data} visualizes some preference data from inference process.
\begin{table}[t] 
\centering
\caption{
Ablation study on key components of \texttt{ClinCoT}. `w/o ClinCoT' removes intermediate reasoning and directly outputs answers. `w/ naive DPO' applies standard DPO loss. `w/o iterative learning' generates all preference pairs in a single pass and trains only once. `w/o $\gamma$ model' sets  $\gamma=0$ during response evaluation stage.  `single evaluator' use a single Med-LLM LLaMA3-Med42-7B as evaluator.}
\resizebox{\linewidth}{!}{
\begin{tabular}{l|cc|cc|ccc}
\toprule
\multirow{2}{*}{Models} & \multicolumn{2}{c|}{\textbf{SLAKE}} & \multicolumn{2}{c|}{\textbf{VQA-RAD}} & \multicolumn{3}{c}{\textbf{IU-Xray}}  \\ 
 & Open & Closed & Open  & Closed & BLEU  & ROUGE-L & METEOR   \\ \midrule
\texttt{ClinCoT} w/o SFT   &     {53.73}\scriptsize{$\pm0.3$} &  {73.41}\scriptsize{$\pm0.2$} &  {33.15}\scriptsize{$\pm0.2$} &  {65.88}\scriptsize{$\pm0.1$} &  {24.96}\scriptsize{$\pm0.2$} &  {29.79}\scriptsize{$\pm0.1$} & {35.44}\scriptsize{$\pm0.2$} \\ \midrule
w/o \texttt{ClinCoT} & 41.09\scriptsize{$\pm0.2$} & 56.83\scriptsize{$\pm0.1$}  & 24.55\scriptsize{$\pm0.1$}  & 51.38\scriptsize{$\pm0.3$} & 17.13\scriptsize{$\pm0.2$} & 23.44\scriptsize{$\pm0.3$} &  26.85\scriptsize{$\pm0.3$} \\
w/ naive DPO   &  50.22\scriptsize{$\pm0.1$}  &   71.39\scriptsize{$\pm0.3$}      &  31.06\scriptsize{$\pm0.4$}  &  62.61\scriptsize{$\pm0.1$}  &  22.76\scriptsize{$\pm0.2$} &  28.51\scriptsize{$\pm0.1$} & 33.18\scriptsize{$\pm0.4$}  \\ 
w/o iterative learning & 49.97\scriptsize{$\pm0.5$}  & 69.83\scriptsize{$\pm0.4$}  &  30.23\scriptsize{$\pm0.4$} & 60.75\scriptsize{$\pm0.3$}   & 21.84\scriptsize{$\pm0.6$}  & 26.19\scriptsize{$\pm0.3$}  & 31.91\scriptsize{$\pm0.5$} \\
w/o $\gamma$ & 46.53\scriptsize{$\pm0.1$} & 65.37\scriptsize{$\pm0.2$} & 28.71\scriptsize{$\pm0.1$} &  57.69\scriptsize{$\pm0.1$} & 20.03\scriptsize{$\pm0.3$}  & 25.42\scriptsize{$\pm0.3$}  &  29.89\scriptsize{$\pm0.1$}  \\
single evaluator &  52.16\scriptsize{$\pm0.7$} &  72.28\scriptsize{$\pm0.4$}  &  32.34\scriptsize{$\pm0.5$} & 64.95\scriptsize{$\pm0.2$}  &  23.62\scriptsize{$\pm0.4$}  &   28.81\scriptsize{$\pm0.3$} &  34.73\scriptsize{$\pm0.2$}   \\
\bottomrule 
\end{tabular}
}
\label{table:ablation}
\end{table}

\subsection{Ablation Study}
In Table \ref{table:ablation}, we present the ablations on key components. \textbf{1) Image-level CoT:} Removing CoT causes a significant drop, confirming its necessity. This highlights the potential of our method and suggests future work for improving region accuracy. \textbf{(2)Margin-aware DPO :} Results shows consistently degrades performance, indicating that incorporating preference score gaps is crucial for distinguishing subtle differences between reasoning chains. \textbf{(3) Iterative learning:} Eliminating iterative updates leads to noticeable declines, suggesting that dynamically generating preference data can help maintain alignment as the model evolves. \textbf{(4) Response evaluation:} Removing the next-step score or using a single evaluator reduces performance, particularly on report generation, showing that stable, consensus-aware scoring improves long-horizon reasoning quality.

\section{Conclusion}
We propose \texttt{ClinCoT} that shifts preference optimization from response-level to hypotheses-driven reasoning. By combining disease-conditioned region proposals, consensus-weighted margin optimization, and iterative learning, ClinCoT aligns intermediate reasoning with localized pathological evidence. Experiments on medical VQA and report generation benchmarks show consistent improvements over strong medical baselines, with further gains under SFT, These results demonstrate that region-level clinical reasoning can be embedded into preference learning to improve factual grounding and reasoning stability in Med-VLMs.
%
%
%
\bibliographystyle{splncs04}
\bibliography{main}

\end{document}